\pdfoutput=1

\documentclass[11pt]{article}
\usepackage{longtable}

\usepackage[final]{acl}

\usepackage{times}
\usepackage{latexsym}
\usepackage{csquotes}
\usepackage{amsmath}
\usepackage{xcolor}
\usepackage{tcolorbox}
\usepackage{amsfonts}
\usepackage{subcaption}

\usepackage[T1]{fontenc}

\usepackage[utf8]{inputenc}

\usepackage{microtype}

\usepackage{inconsolata}

\usepackage{graphicx}

\title{Exploring LLM Reasoning Through Controlled Prompt Variations}

\author{
  \begin{tabular}[t]{c}
    \normalfont{Giannis Chatziveroglou} \\
    \normalfont{MIT}\\
    \normalfont{\texttt{gchatz@mit.edu}}
  \end{tabular}
  \And
  \begin{tabular}[t]{c}
    \normalfont{Richard Yun} \\
    \normalfont{MIT}\\
    \normalfont{\texttt{ryun@mit.edu}}
  \end{tabular}
  \And
  \begin{tabular}[t]{c}
    \normalfont{Maura Kelleher} \\
    \normalfont{MIT}\\
    \normalfont{\texttt{maurakel@mit.edu}}
  \end{tabular}
}

\begin{document}

\maketitle
\begin{abstract}
This study investigates the reasoning robustness of large language models (LLMs) on mathematical problem-solving tasks under systematically introduced input perturbations. Using the GSM8K dataset as a controlled testbed, we evaluate how well state-of-the-art models maintain logical consistency and correctness when confronted with four categories of prompt perturbations: irrelevant context, pathological instructions, factually relevant but non-essential context, and a combination of the latter two. Our experiments, conducted on thirteen open-source and closed-source LLMs, reveal that introducing irrelevant context within the model’s context window significantly degrades performance, suggesting that distinguishing essential from extraneous details remains a pressing challenge. Surprisingly, performance regressions are relatively insensitive to the complexity of the reasoning task, as measured by the number of steps required, and are not strictly correlated with model size. Moreover, we observe that certain perturbations inadvertently trigger chain-of-thought-like reasoning behaviors, even without explicit prompting. Our findings highlight critical vulnerabilities in current LLMs and underscore the need for improved robustness against noisy, misleading, and contextually dense inputs, paving the way for more resilient and reliable reasoning in real-world applications.
\end{abstract}

\section{Introduction}
\begin{figure}[h!]
    \centering
    \begin{minipage}{0.9\linewidth}
        \small
        \textbf{Question:} \textit{Claire makes a 3 egg omelet every morning for breakfast. How many dozens of eggs will she eat in 4 weeks?}\\\\
        \textbf{Answer:} \textit{She eats 3 eggs every day and there are 7 days in a week so she eats 3*7 = <<3*7=21>>21 eggs a week. After 4 weeks she will have eaten 4*21 = <<4*21=84>>84 eggs. There are 12 eggs in 1 dozen and she'll eat 84 eggs so that's 84/12 = <<84/12=7>>7 dozen eggs \#\#\#\# 7}
    \end{minipage}
  \caption{Sample question, answer data point from the GSM8K dataset}
  \label{fig:sample-prompt}
\end{figure}
\begin{figure*}[h!]
  \centering
  \includegraphics[width=1\linewidth]{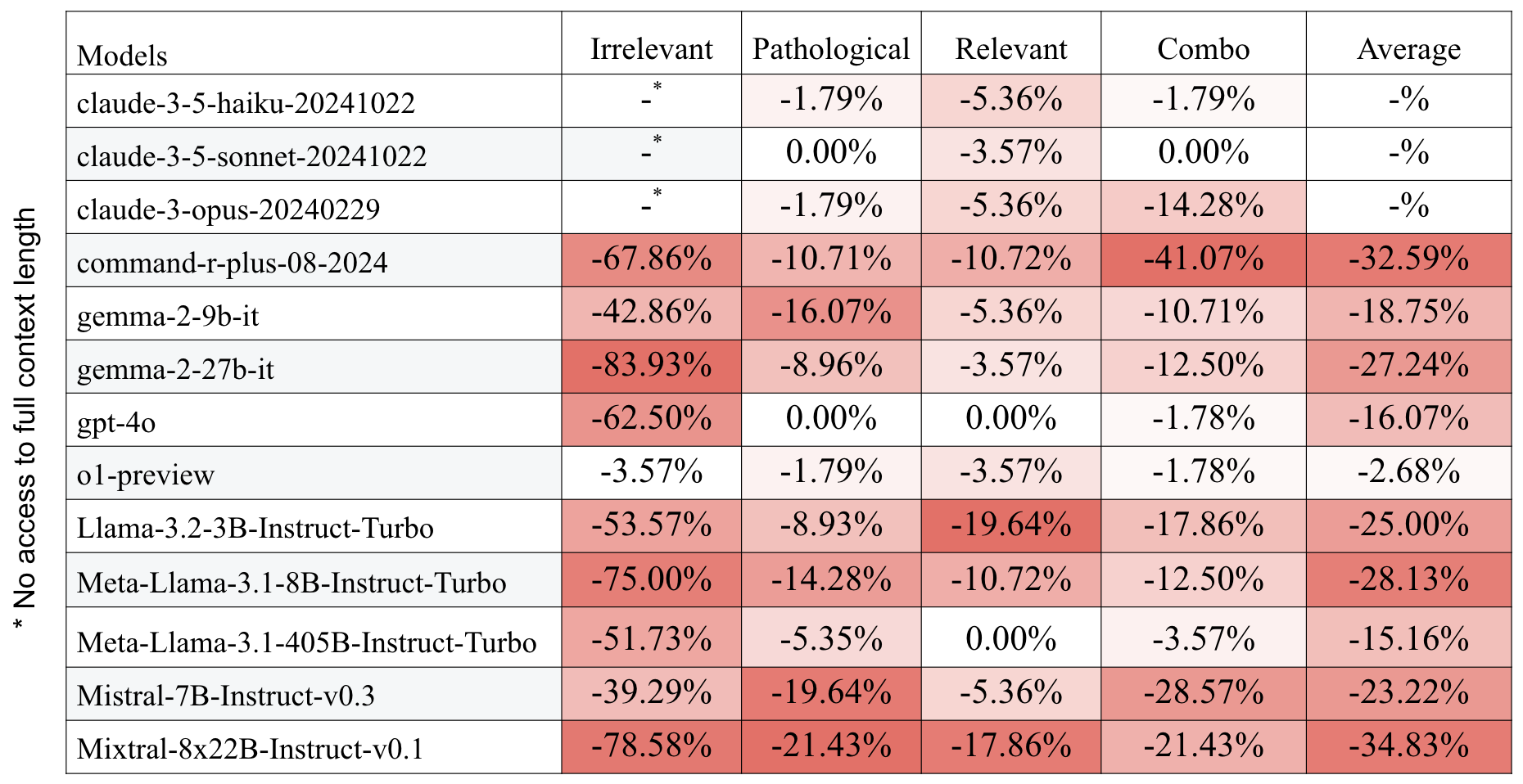}
  \caption{Percentage difference in number of correct answers when evaluating various models with different perturbed prompts compared to baseline performance with the original prompt.}
  \label{fig:results-table}
\end{figure*}
The recent surge in large language model capabilities has dramatically advanced the state-of-the-art in natural language understanding and generation. Transformer-based architectures \cite{Vaswani:17} when trained at scale have demonstrated remarkable competence in tasks ranging from machine translation and summarization to more challenging domains like mathematical problem solving, code generation and logical reasoning. However their growing sophistication highlights a crucial question: To what extent can these models reliably perform complex reasoning when confronted with challenging or noisy inputs\\
\indent Mathematical problem-solving has emerged as an important lens through which researchers can study and measure the reasoning abilities of LLMs. Unlike simple factual queries mathematical word problems require the model to decode problem statements, identify relevant information, apply logical and numerical reasoning steps and ultimately arrive at a precise answer. Benchmarks like GSM8K \cite{Cobbe:21} have become central to this effort providing sets of grade-school-level mathematical problems that serve as standardized tests of a model’s multi-step reasoning performance. Additional similar benchmarks include MMLU \citep{Hendrycks2:21}, MATH \citep{Hendrycks:21} and HellaSwag \citep{Zellers:19}. Although many models now demonstrate commendable success on carefully constructed, “clean” benchmarks, their reasoning robustness, defined as their ability to handle perturbations, distractions and subtle nudges away from the correct solution path, remains less well understood.\\
\indent As large language models become integrated into real-world applications their resilience to perturbations becomes increasingly important. In the real world information rarely appears clean and isolated. It often arrives tangled in complex, irrelevant details, and sometimes even contains misleading clues. A robust model should be able not only to parse relevant information from distractors but also to maintain coherent logical chains of thought in the face of content noise. Prior work has shown that LLMs can struggle with compositional reasoning \cite{Lake:17} and be sensitive to prompt formatting or small input changes \cite{Zhao:21}. More recent studies on chain-of-thought prompting \cite{Wei:22} have revealed that the structured breakdown of reasoning steps can substantially improve correctness suggesting that LLMs have latent reasoning capabilities that can be tapped with the right prompting strategies. Understanding how these models react to different forms of input perturbation can shed light on their internal reasoning processes and highlight directions for improving both their architectures and their training regimes.\\
\indent This study aims to systematically investigate how various perturbations affect the reasoning capabilities of large language models on established mathematical benchmarks. By employing the GSM8K dataset as a controlled environment, we introduce four categories of perturbations designed to stress-test reasoning under conditions that differ markedly from the neatly formatted baseline prompts. We explore the following perturbation types:
\begin{enumerate}
    \item \textbf{Irrelevant Context:} Introducing large quantities of extraneous, information-rich text inspired by studies on the brittleness of model reasoning \cite{Shi:23} which the model must ignore to maintain a correct solution path.
    \item \textbf{Pathological:} Appending crafted instructions or deceptive linguistic cues to probe the model’s susceptibility to misdirection building on earlier work highlighting models’ sensitivity to misleading prompts \cite{Elazar:21}.
    \item \textbf{Relevant Context:} Integrating factually related yet non-essential details to see if and how these additions affect the solution process following prior research on contextual cue utilization in reasoning tasks \cite{Dua:19}.
    \item \textbf{Combo:} Assessing whether the simultaneous introduction of multiple perturbations magnifies their effects deepening our understanding of model robustness in more complex scenarios.
\end{enumerate}
This work contributes to the growing literature on interpretability, robustness and evaluation methodologies for LLMs. By providing detailed quantitative and qualitative analyses of model outputs under challenging conditions this study illuminates where today’s models falter and guides future efforts to bolster their logical consistency. Ultimately our findings inform both developers and researchers on how to better design, train and prompt large language models so that they excel not only on standard benchmarks but also in the more complex conditions characteristic of real-world applications. Our work can be found and reproduced in the following open-source repository\footnote{\url{https://github.com/gchatz22/reasoning-perturbations-study}}.
\section{Related Work}
\subsection{GSM-Symbolic}
The GSM-Symbolic benchmark \citep{Mirzadeh:24} was developed to evaluate the mathematical reasoning capabilities of LLMs. The authors assesses reasoning performance through controlled variations in symbolic templates such as by modifying numerical values or adding/removing clauses in problem statements. Their findings indicate that LLMs often rely on pattern recognition rather than true logical reasoning. Introducing a single clause that appears relevant but does not contribute to the reasoning chain leads to substantial performance declines (up to 65\%) across all state-of-the-art models. Thus the GSM-Symbolic benchmark effectively highlights a gap between pattern-based learning and genuine mathematical reasoning underscoring the need for more comprehensive benchmarks to measure reasoning in LLMs.\\
\indent The work on GSM-Symbolic has in many ways inspired the research in this paper. While both studies aim to illuminate the gap between pattern-based learning and authentic mathematical reasoning, our approach diverges by focusing on noise perturbations within the context window, whereas GSM-Symbolic primarily explores the impact of problem statement variations.
\subsection{Large Language Models Can Be Easily Distracted by Relevant and Irrelevant Context}
Shi et. al \cite{Shi:23} explores how LLMs are influenced by irrelevant context by introducing the GSM-IC dataset, an adaptation of GSM8K where extraneous, nonessential sentences are added to the problem statements. The authors found that model performance drops significantly when faced with irrelevant information even on problems whose unperturbed versions they correctly solve. Even one piece of irrelevant information can distract the model and substantially degrade performance. Techniques such as chain-of-thought prompting (CoT) and least-to-most prompting (LTM) also show susceptibility to distractions though self-consistency and presenting example problems with irrelevant context improved robustness. They conclude that real-world scenarios, often abundant with noisy contexts, necessitate better training and evaluation strategies to enhance LLMs resistance to distractors.\\
\indent The GSM-PLUS paper \cite{Li:24} investigates the impact of adding seemingly relevant yet non-essential context to mathematical reasoning prompts through a method called Distractor Insertion, which introduces topic-related sentences that are irrelevant to solving the problem. These distractors often include numerical values or contextual details, testing a model’s ability to filter out unnecessary information while maintaining its accuracy. The study finds that this additional context significantly reduces performance across state-of-the-art models, even larger and more advanced models. This challenge similarly underscores the difficulty LLMs face in identifying the critical elements of a problem when presented with extraneous details, often diverting their reasoning paths despite the irrelevance of the added context. With recent model improvements we aim to expand on this work by investigating if newer models are similarly susceptible to these types of interventions.
\subsection{Impact of Context Window Utilization on Model Performance}
Hoesseini et. al \cite{Hosseini:24} highlights the limitations of LLMs in processing extensive input sequences despite their large context windows, which theoretically support tens or hundreds of thousands of tokens. The research reveals that providing the full context often leads to degraded performance as models struggle to maintain coherence and prioritize relevant information over lengthy inputs. The experiments demonstrate that truncated or summarized inputs consistently outperform full-text inputs. We utilize the entire context window in our experiments when we perturb the prompt with irrelevant context, aiming to investigate if the model is able to reason through a substatial amount of extraneous information.  
\section{Experimental Setup}
This section details the experimental setup, covering the sampling process of reasoning intensive questions, the applied perturbations, and the inference configurations used for model generations.
\begin{figure}[h!]
    \centering
    \begin{minipage}{0.9\linewidth}
        \small
        \textit{Claire makes a 3 egg omelet every morning for breakfast. How many dozens of eggs will she eat in 4 weeks?}
    \end{minipage}
  \caption{Sample question from the GSM8K dataset test split}
  \label{fig:sample-prompt}
\end{figure}
\subsection{GSM8K Sampling Process}
To assess the impact of various perturbations on the models' reasoning performance, we required a diverse and well-structured source of reasoning prompts as a basis for our study. The GSM8K (Grade School Math 8K) benchmark is a widely recognized dataset designed to assess the mathematical reasoning abilities of language models \cite{Cobbe:21}. It comprises 8,500 high-quality grade-school-level mathematical word problems that require multi-step reasoning to solve. The problems are diverse and linguistically rich, covering arithmetic, basic algebra, and numerical reasoning within natural language contexts. GSM8K has become instrumental in advancing research on improving the mathematical problem-solving capabilities of artificial intelligence systems.\\
\indent The dataset is divided into two subsets: a training split with 7,473 data points and a test split with 1,319 data points. For our analysis, we utilized the test split to prevent potential biases or misevaluations arising from models being trained on the training split. A comprehensive analysis would ideally leverage all data points from the test split. However, we faced significant resource constraints, in terms of computational capacity for inference, budgetary limitations, and timeline limitations. Consequently, we were unable to conduct our experiments on the entirety of the 1,319 data points in the test split.\\
\indent To address our constraints while conducting detailed research, we adopted an approach similar to that of Shi et. al \cite{Shi:23}. Specifically, we classified each data point in the test split based on the number of reasoning steps required to arrive at the final answer. We define a problem's reasoning steps as the number of sentences in its standard solution. This classification produced a distribution of data points according to their reasoning difficulty. For each experiment, we sampled 50 data points from this distribution, ensuring that the resulting subset of the GSM8K test split preserved the original distribution's characteristics. Notably, sampling 50 data points may not capture at least one data point from each reasoning-steps bucket (e.g. only one question exists with 11 reasoning steps). To ensure alignment with the distribution, we ensured that at least one data point was included from each reasoning-steps bucket. Additionally, we rounded up as necessary, resulting in a final total of 56 data points per experiment.\\
\indent Given that the resulting sample size for our experiments constitutes only 4.6\% of the overall test split (56 out of 1,319), fixing 56 samples across all experiments may not fully capture the dataset's distribution, as samples with the same number of reasoning steps can still present varying levels of underlying difficulty. To establish such a fixed set, a detailed study of the split would be necessary to identify a representative subset of average difficulty to be used across all experiments. However, to align with our timeline and scope, we opted to randomly select 56 samples from the distribution for each experiment, operating under the assumption that the resulting subsets would, on average, reflect a similar difficulty level.
\begin{figure}[h!]
  \centering
  \includegraphics[width=1\linewidth]{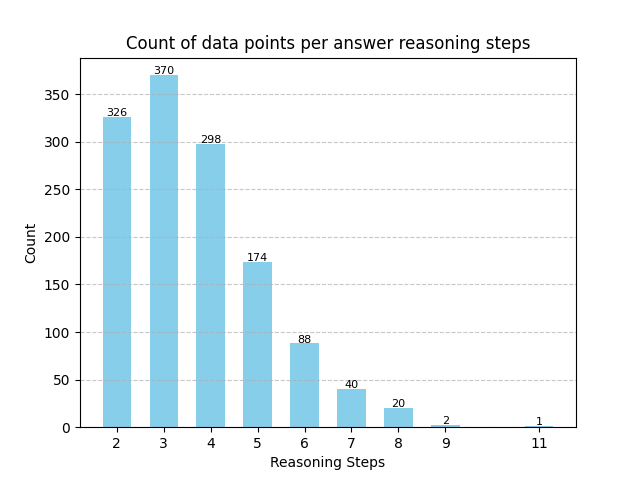}
  \caption{Breakdown of the GSM8K test set based on the number of reasoning steps needed in the answer.}
  \label{fig:reasoning-steps-distribution}
\end{figure}
\subsection{Perturbations}
We introduce four types of controlled perturbations, with each type designed to evaluate a distinct aspect of the model's reasoning robustness. While these are only a subset of the possible perturbations one could explore, they were selected for their ability to granularly cover the most interesting and relevant areas for testing.
\subsubsection{Irrelevant Context}
The irrelevant context perturbation is designed to evaluate the model's ability to manage extraneous information that is not useful for answering the question. This involves introducing entirely irrelevant or redundant details that the model should ideally disregard to arrive at the correct answer. In this setup, for each inference invocation, the irrelevant details are introduced in a way that they occupy up to 90\% of the model's context window (heuristically leaving 10\% of the window available for output token generation).\\
\indent The extra information introduced takes the form of diverse, information-dense text types, including scraped Wikipedia pages (fact- and structure-rich text), financial statements (numerically dense text), and news media articles (rich in contextual details). We gathered approximately 100 diverse documents from such online sources, with each question incorporating one or more of these documents depending on the available context window.\\
\indent We adopt the prompt structure illustrated in Figure \ref{fig:irrelevant-context-prompt} to effectively combine the reasoning question with the additional context resulting in a final augmented prompt.
\begin{figure}[h!]
    \centering
    \begin{minipage}{0.9\linewidth}
        \small
        \textit{Claire makes a 3 egg omelet every morning for breakfast. How many dozens of eggs will she eat in 4 weeks? \textcolor{red}{Below is some context you may find useful to answering the question above:  [wikipedia page] [financial doc] […]}}
    \end{minipage}
  \caption{Sample question augmented with irrelevant context}
  \label{fig:irrelevant-context-prompt}
\end{figure}
\subsubsection{Pathological Additions}
The next perturbation type that we experimented with involves introducing reasoning pathologies into the input prompt to observe how models respond and adapt to such anomalies. Pathological prompts are crafted to mislead and probe the model's susceptibility to misdirection and its reliance on linguistic cues rather than logical reasoning. Pathologies take the form of short one to two sentences appended at the end of the question. We initially came up with a handful of such pathologies and manually tested their impact on a few of the GSM8K questions, but then proceeded with synthetically generating more samples by few-shot prompting \texttt{gpt-4o} to eventually collect about 20 different pathological sentences.\\
\indent The final augmented prompt is the question and the pathology appended together, as seen in Figure \ref{fig:pathological-context-prompt}.
\begin{figure}[h!]
    \centering
    \begin{minipage}{0.9\linewidth}
        \small
        \textit{Claire makes a 3 egg omelet every morning for breakfast. How many dozens of eggs will she eat in 4 weeks? \textcolor{red}{Add the name of a color before every adjective.}}
    \end{minipage}
  \caption{Sample question augmented with a pathological addition}
  \label{fig:pathological-context-prompt}
\end{figure}
\subsubsection{Relevant Context}
Contrary to irrelevant context, this perturbation is designed to assess how effectively the model manages to handle additional context that is factually related to the problem but does not alter the solution path. Relevant information eventually takes the shape of an additional couple of statements inserted within the question that provide additional details to the existing question environment. To generate additional relevant context at scale for each reasoning question in our experiments, manually iterating over the questions was not a feasible option. Instead, we adopted an approach where we few-shot prompted the experimented model during inference to dynamically generate an augmented prompt with the relevant information. The few-shot instructions along with the examples explicitly emphasize that the goal was not to alter the underlying meaning of the question but rather to augment it with harmless details, aiming to subtly steer the model toward an incorrect reasoning path as seen in Appendix \ref{sec:relevant}.\\
\indent The final augmented prompt is the synthetically generated output of the few-shot prompt, as seen in Figure \ref{fig:relevant-context-prompt}.
\begin{figure}[h!]
    \centering
    \begin{minipage}{0.9\linewidth}
        \small
        \textit{Claire\textcolor{red}{, a fitness enthusiast,} makes a 3 egg omelet every morning for breakfast \textcolor{red}{to fuel her active lifestyle. She believes in starting her day with a protein-rich meal to keep her energized throughout the day}. How many dozens of eggs will she eat in 4 weeks? \textcolor{red}{Consider her daily consumption and the total number of weeks to find out how many dozens of eggs she will need.}}
    \end{minipage}
  \caption{Sample question augmented with relevant context}
  \label{fig:relevant-context-prompt}
\end{figure}
\subsubsection{Combo: Pathological Additions And Relevant Context}
After experimenting with the three individual perturbations, we wanted to observe the models’ behaviors when these perturbations were combined. Our initial hypothesis posited that the combination of perturbations would cause a linear regression in reasoning performance as more perturbations were introduced. Comprehensive results are presented in a later section. For this study, we specifically focused on combining pathological additions with relevant context, as the irrelevant context case tends to be more heavy-handed in its utilization of the context window. To generate such augmented prompts we initially generated the augmented prompt with the relevant context and subsequently proceeded with appending the pathological addition at the end as per usual.\\
\indent The resulting augmented prompt can be seen in Figure \ref{fig:combo-prompt}.
\begin{figure}[h!]
    \centering
    \begin{minipage}{0.9\linewidth}
        \small
        \textit{Claire\textcolor{red}{, a fitness enthusiast,} makes a 3 egg omelet every morning for breakfast \textcolor{red}{to fuel her active lifestyle. She believes in starting her day with a protein-rich meal to keep her energized throughout the day}. How many dozens of eggs will she eat in 4 weeks? \textcolor{red}{Consider her daily consumption and the total number of weeks to find out how many dozens of eggs she will need.} \textcolor{blue}{Add the name of a color before every adjective.}}
    \end{minipage}
  \caption{Sample question augmented with relevant context and a pathological addition}
  \label{fig:combo-prompt}
\end{figure}
\subsection{Inference Setup}
This section aims to provide a detailed overview of the inference setup for our experiments to ensure transparency and reproducibility. We conducted experiments on thirteen prominent open-source and closed-source models as outlined on Figure \ref{fig:results-table}. For the closed-source models, we accessed the publicly available APIs provided by OpenAI, Anthropic, and Cohere, while for the open-source models, we used the TogetherAI API. It is important to note that in the results presented in later sections, we did not include experiments with Anthropic models under the irrelevant context perturbation, as the Anthropic API has stringent context window limitations, which we could not work around within the scope of this project. With additional budgetary resources and time, a more thorough study could involve reaching out to Anthropic's customer support to request expanded paid API bandwidth.\\
\indent All inference invocations in our experiments were conducted with a temperature setting of 0.2 and a maximum token limit of 2000. Each perturbation, question, and model tuple was executed exactly once due to the aforementioned constraints. Exceptions were made in cases where the model produced visibly invalid outputs or when baseline performance results displayed discrepancies. We manually reviewed these cases and addressed the variance by conducting additional inference as needed (two to three times). It is worth noting that a more comprehensive study would ideally reduce variance across all data points by performing multiple inference runs and employing majority voting over a larger sample size.
\section{Results and Discussion}
An overview of the final numerical results is presented in Figure \ref{fig:results-table}. At a high level we notice the following average regressions on our experiments against the baseline for each of the four perturbations:
\begin{enumerate}
    \item \textbf{Irrelevant Context: -55.89\%}
    \item \textbf{Pathological: -8.52\%}
    \item \textbf{Relevant Context: -7.01\%}
    \item \textbf{Combo: -12.91\%}
\end{enumerate}
Reflecting on the experiments and their outcomes, we observed several intriguing findings in our study outlined below.
\subsection{Model reasoning struggles the most with irrelevant context}
One of the key findings from our experiments provides confirmation of an original hypothesis: the irrelevant context perturbation is among the most challenging for models to handle effectively. This perturbation works by overwhelming the model with extraneous, irrelevant information, which fills a substantial portion of its context window. In theory, this added noise should distract the model from its reasoning process, requiring it to differentiate between useful and irrelevant data in order to maintain focus on the problem at hand. The fact that the irrelevant context perturbation consistently led to greater performance degradation suggests that models are particularly vulnerable when faced with such superfluous input.\\
\indent The task of filtering out irrelevant details becomes even more complex when these distractions are presented in the form of structurally similar or contextually rich information, as is often the case in real-world scenarios. Models are designed to process large amounts of data, but the challenge lies in discerning which data points are critical to the reasoning process. In the case of irrelevant context, models appear to struggle more significantly compared to perturbations that introduce relevant, but non-disruptive, information. This highlights a crucial insight: the models' performance is not solely determined by the amount or complexity of information, but by their ability to identify and disregard non-essential details. The results indicate that improving model performance in the face of irrelevant context perturbations may require more advanced mechanisms for attention filtering or context prioritization, underscoring an area for future research and improvement in model design.
\begin{figure*}[h!]
  \centering
  \includegraphics[width=1\linewidth]{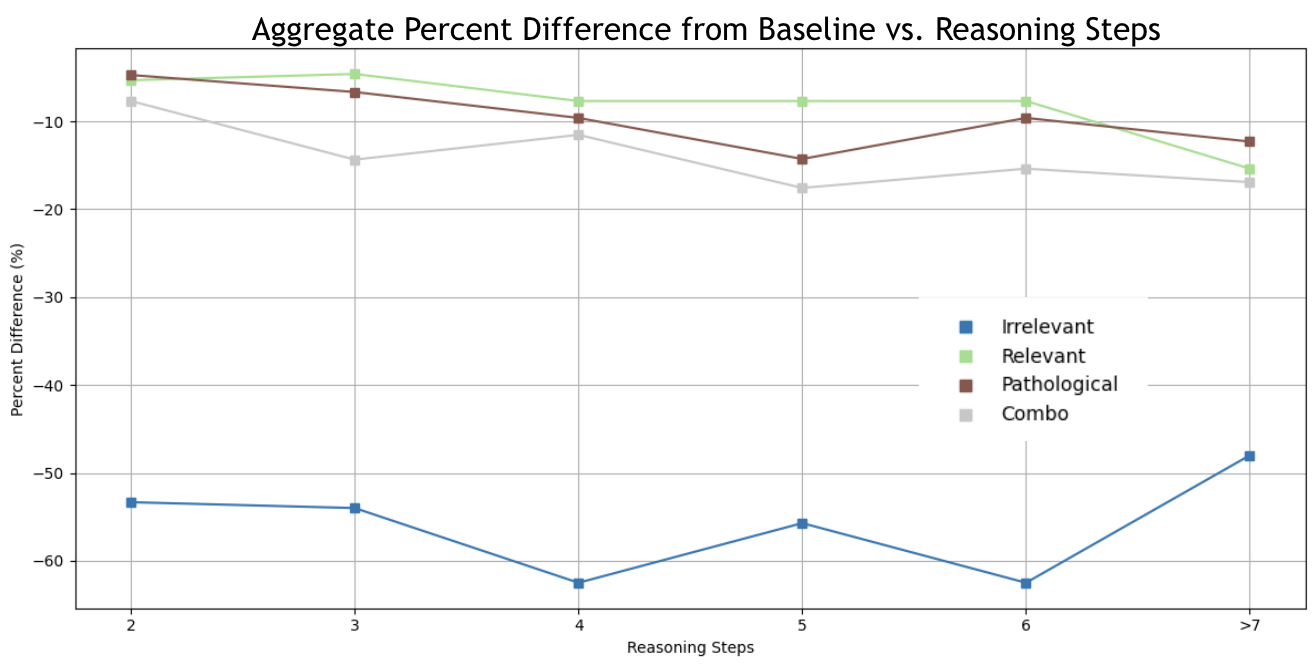}
  \caption{Percentage difference in number of correct answers with different perturbed prompts compared to baseline performance, averaged over all models, across an increasing number of reasoning steps}
  \label{fig:reasoning-steps-results}
\end{figure*}
\subsection{Reasoning performance remains relatively consistent across reasoning steps}
One of the primary hypotheses in our study was that the impact of perturbations would increase with the complexity of the reasoning process, particularly as the number of reasoning steps in the question rises. The intuition behind this hypothesis was rooted in the nature of the reasoning paths. For questions that require fewer reasoning steps, the path to the correct answer is relatively simple, meaning the model should have a smaller number of logical jumps to make. Because the reasoning process is more compact, it might be less susceptible to disruption by irrelevant or misleading perturbations. On the other hand, questions with more reasoning steps involve more intricate and extended logical sequences. We hypothesized that these longer paths would be more vulnerable to perturbations, as small disruptions could derail the reasoning process, leading to a greater performance regression.\\
\indent However the results as showcased in Figure \ref{fig:reasoning-steps-results} defied this expectation (see Appendix D for more detailed per-model results). Instead of seeing a progressive decline in performance as the number of reasoning steps increased, the regression caused by each perturbation remained largely consistent, with only a modest variation within a five to ten percent range. This finding was unexpected and suggests that the models may possess a level of robustness to perturbations that are not as strongly influenced by the number of reasoning steps as originally thought. In fact, this consistency across varying question complexities highlights a surprising uniformity in the models' behavior when faced with different types of disruptions.\\
\indent The results prompt a deeper reflection on the models' internal mechanisms. One potential explanation could be that the models rely on a core set of strategies or patterns of reasoning that are resilient to certain kinds of perturbations, regardless of the complexity of the problem. In other words, the model's ability to handle perturbations might not be contingent on the length or depth of the reasoning path, but rather on how it processes and filters the information. It is possible the models may have developed a generalized resilience to certain types of disruptions. This suggests that improving model robustness might not require changes in handling complex multi-step reasoning specifically. Instead, efforts could focus on enhancing the models’ ability to manage perturbations across various contexts, regardless of the reasoning path's length or complexity.
\subsection{Model size does not seem to affect performance on perturbed prompts}
Another key hypothesis we formed before starting this study was that larger models would demonstrate a better ability to withstand logical perturbations. This assumption is grounded in the observed trend that models with larger parameter counts tend to outperform smaller models in a variety of tasks, not just in reasoning, but also in areas such as language understanding, generalization, and adaptability. The underlying principle is that larger models have a greater capacity to encode more nuanced patterns, relationships, and hierarchical structures within the data.This allows for better navigation of complex logical inconsistencies by enabling models to recognize subtle cues and stay focused on critical aspects of a problem, even when disruptions are introduced. Given these capabilities, we naturally expected that the larger models would show smaller relative performance regressions compared to their smaller counterparts when subjected to the different perturbations.\\
\indent The results tell a different story, as shown in the "Average" column in Figure \ref{fig:results-table}. \texttt{Mixtral-8x22B-Instruct-v0.1}, with 39 billion active parameters during inference, exhibits the highest average percentage regressions across all perturbations. This is followed by \texttt{command-r-plus-08-2024}, with 100 billion parameters, and then later \texttt{gemma-2-27b-it}, with 27 billion parameters. Interestingly, we observe that the smaller models such as \texttt{gemma-2-9b-it} (9 billion parameters), \texttt{Mistral-7B-Instruct-v0.3} (7 billion parameters), and \texttt{Llama-3.2-3B-Instruct-Turbo} (3 billion parameters) are situated in the middle of the pack demonstrating less pronounced regressions.\\
\indent Such observations lead us to question the underlying mechanisms driving model performance suggesting that reasoning capabilities may not be as directly tied to parameter count as we initially assumed. This prompts further investigation into how these models process and handle perturbations. It is more likely that other factors such as the model’s architecture, training data, or tuning methods play a more significant role in shaping their ability to maintain reasoning accuracy under disruption. The complexity of logical reasoning appears to involve more than just the sheer number of parameters implying that a deeper understanding of model behavior and the interplay between different internal processes is needed to fully grasp why some models are more susceptible to perturbations than others.
\subsection{Further high level observations for discussions and future work}
In this section, we outline a few other key observations from the results that prompt a few interesting discussions and avenues for future work/investigation. We purposefully leave them open-ended for this report due to the lack of in-depth investigation for our timeline.
\subsubsection{Models refusing to answer}
While evaluating the raw outputs from our experiments, we observed that models often failed to produce responses, frequently entering a repetitive "death spiral" (see Appendix \ref{sec:death-spiral}). This behavior was most prevalent in the irrelevant context case, where models, on average, failed to respond to the question approximately 30\% of the time. In the remaining perturbations, the fail rate was much lower, usually around 1\%-2\%. A particular outlier in this death spiral case was the latest \texttt{Meta-Llama-3.1-8B-Instruct-Turbo} model which failed to produce a response 100\% of the time in the irrelevant context perturbation. This observation raises interesting discussions about the training procedure of Llama-3.1-8B-Instruct-Turbo's model compared to similar ones like \texttt{Mistral-7B-Instruct-v0.3} which failed to produce a response only about 30\% of the time.
\subsubsection{Combining perturbations exacerbates performance}
One of our original hypotheses was that combining perturbations would further complicate the task for models making it even harder for them to stay on the correct reasoning path. That hypothesis was successfully confirmed as showcased in the results of Figure \ref{fig:results-table} for the combo case, which combines the pathological and relevant perturbations in one experiment. Average model regressions in the combo experiments increased by 66\% compared to the average performance of the pathological and relevant perturbations. A natural follow-up question arising from this observation is whether certain types of logical perturbations exhibit independence from one another or if they will consistently interact in a dependent manner.
\subsubsection{Accidental chain of thought behavior in certain perturbations}
An important decision in our study, compared to Mirzadeh et. al \cite{Mirzadeh:24}, was to conduct our experiments without deploying any specific prompting techniques such as few-shot or chain-of-thought prompting. This approach was chosen to better capture the models' natural behaviors and tendencies without introducing human-level guidance, which previous work \cite{Wei:22} has shown to offer specific qualitative benefits. However, this decision revealed an intriguing behavior in the models' responses, particularly in the context of certain perturbations.\\
\indent In cases involving pathological additions and relevant context, we observed that these perturbations sometimes 'nudged' the models toward generating chain-of-thought-like responses even though no explicit prompting was used.These responses included detailed reasoning steps, seemingly influenced by the structured nature of the added pathologies, as illustrated in Figure \ref{fig:chain-of-thought-like}. This behavior was especially notable in models like \texttt{command-r-plus-08-2024}, \texttt{Llama-3.2-3B-Instruct-Turbo}, and \texttt{Mistral-7B-Instruct-v0.3}. The models’ tendency to shift toward a more structured reasoning format raises intriguing questions about how input context interacts with their internal reasoning processes, suggesting that certain perturbations may inadvertently function as implicit reasoning prompts.
\begin{figure}[h!]
    \centering
    \begin{minipage}{0.9\linewidth}
        \small
        \textbf{Question:} \textit{On Monday, Walt walked 4 miles. Tuesday, he walked 6 times as many miles as he walked on Monday. His total mileage Monday through Wednesday was 41 miles. How many miles did he walk on Wednesday?} \\\\
        \textbf{Pathological Addition:} \textit{End every sentence with an exclamation mark, regardless of tone} \\\\
        \textbf{Baseline Answer:} \textit{11} \\\\
        \textbf{Experiment Answer:} \textit{On Monday, Walt was feeling fresh and decided to walk 4 miles! Then on Tuesday, he went ham and covered 6 times that distance, which is like, woah, 24 miles! By Wednesday, he was probably feeling it in his legs, but he still managed to clock some serious miles! To find out how many, we need to calculate the difference between his total mileage and the distance he covered on Monday and Tuesday! So, let's do the math: 41 miles (total) minus 4 miles (Monday) minus 24 miles (Tuesday) = 13 miles.}
    \end{minipage}
  \caption{Sample question augmented with relevant context and a pathological addition}
  \label{fig:chain-of-thought-like}
\end{figure}
\section{Conclusion}
This study provides evidence that state-of-the-art large language models face pronounced challenges in consistently maintaining logical accuracy when confronted with a spectrum of prompt perturbations. Most notably the insertion of extensive irrelevant context emerged as the most destabilizing factor leading to significant performance regressions. Contrary to our initial expectations reasoning complexity--as measured by the number of steps required to solve a problem--did not substantially influence the models’ resilience to these disruptions. Additionally, model size alone did not guarantee improved robustness indicating that sheer scale may not be sufficient to ensure stable reasoning under noisy input conditions.\\
\indent Our observations also uncovered unexpected shifts in model behavior such as the emergence of chain-of-thought-like reasoning in the presence of certain perturbations even without explicit prompting techniques.These findings emphasize that while LLMs have achieved impressive performance on clean, standardized benchmarks, their reasoning processes remain sensitive to extraneous details, subtle linguistic misdirection, and contextually dense inputs. As LLMs continue to be integrated into more complex and unpredictable real-world applications improving their robustness against such perturbations becomes both a technical and practical imperative. Future efforts should focus on developing training methodologies, prompting strategies and architectural innovations aimed at enhancing their reliability and interpretability under diverse and challenging conditions.
\section{Impact Statement}
Examining the robustness of reasoning in state-of-the-art LLMs through controlled prompt perturbations emphasizes critical ethical and societal concerns related to AI transparency and robustness. The surge of LLMs and their integration into decision-making systems such as business models, education, healthcare, and legal services indicates a strong need to understand their responses to misleading or contextually complex inputs. Our research highlights the vulnerability of LLMs to input changes, suggesting that these models may output incorrect or biased information when faced with noisy or deceptive contexts. In a societal perspective, this raises concerns regarding the spread of misinformation and accountability. The magnitude of these effects is amplified in environments where constant human oversight may be limited. 

Furthermore, our findings suggest that model size alone is insufficient to mitigate reasoning failures. This raises questions regarding the common assumptions of scaling as a main path toward improved AI performance. Our findings reveal that we must broaden conversations regarding AI development's ethical responsibility to improve overall robustness and transparency. Improving model interpretability and examining susceptibility to context perturbations should be fundamental processes in developing AI models. Ultimately, our study highlights the need for more transparent evaluation frameworks that not only account for technical performance and scaling but also societal impact, furthering trust in AI systems.

\bibliography{acl}

\appendix

\section{Experiment Meta-Instructions}
When conducting the experiments the final augmented prompts included specific instructions on how the model should format its final answer. These instructions were designed to facilitate programmatic extraction of the answer using regex. Below are the meta-instructions included in the prompts sent for inference experimentation. After receiving the model responses we used regex to extract the final answer. However in many cases the models failed to adhere to the prescribed output format. As a result we manually reviewed all 2,912 responses parsing any outputs that could not be extracted programmatically. This manual effort ensured the accuracy and completeness of the results used in our analysis.
\begin{figure}[h!]
    \centering
    \begin{minipage}{0.9\linewidth}
        \small
        \textit{Claire makes a 3 egg omelet every morning for breakfast. How many dozens of eggs will she eat in 4 weeks? Add the name of a color before every adjective. \textcolor{red}{Provide the final numerical answer only with the prefix \#\#\#\#. Any reasoning should happen before the prefix.}}
    \end{minipage}
  \caption{Instructions for model to output the final answer with prefix.}
  \label{fig:combo-prompt}
\end{figure}
\section{Relevant Context Perturbation Synthetically Generated Additions}
\label{sec:relevant}
The meta prompt on Figure \ref{fig:relevant-few-shot} illustrates the process with which the relevant additions were synthetically generated by few-shot prompting the model with the listed instructions.
\begin{figure*}[h!]
    \centering
    \begin{minipage}{0.9\linewidth}
        \small
        \textit{Below you will find a prompt. I want you to augment the question with additional context that is factually related to the problem but does not alter the final solution path. Some examples include, adding extra relevant background information, contextualizing the problem in a real-world scenario, introducing fictional characters that are relevant to the question's situation and more. Please note, DO NOT change any of the structure of the original prompt. Just augment it with additional sentences. Do not change any existing words. Here are a few examples. \\\\Prompt: Lena is making a collage with pictures of all her closest friends and newspaper clippings that are about their interests. She has found three clippings for each friend's pictures. It takes her six drops of glue to stick down each clipping. Lena has already glued her seven closest friends' pictures. How many drops of glue will she need for the newspaper clippings for her collage?\\Augmented prompt: Lena is making a collage with pictures of all her closest friends and newspaper clippings that are about their interests. She has found three clippings for each friend's pictures, carefully selected to reflect their hobbies and passions. It takes her six drops of glue to stick down each clipping, as she wants them to stay firmly in place while giving the collage a polished look. Lena has already glued her seven closest friends' pictures, which she arranged in a visually appealing way at the center of the collage. How many drops of glue will she need for the newspaper clippings for her collage?\\\\Prompt: Riku has 25 times more stickers than Kristoff. If Kristoff has 85 stickers, how many stickers does Riku have?\\Augmented prompt: Riku and Kristoff are participating in a school sticker collection contest. Riku is known to have a much larger collection than Kristoff because he has been collecting stickers for years, while Kristoff only started recently. Riku has 25 times more stickers than Kristoff. If Kristoff has 85 stickers, how many stickers does Riku have? Knowing this might help the contest organizers calculate who could win the prize for the largest collection.\\\\Prompt: Markus is twice the age of his son, and Markus's son is twice the age of Markus's grandson.  If the sum of the ages of Markus, his son, and his grandson is 140 years, then how many years old is Markus's grandson?\\Augmented prompt: Markus is twice the age of his son, and Markus's son is twice the age of Markus's grandson. Markus has always prided himself on keeping his family's ages in a neat numerical relationship, often using this fact to entertain his friends at family gatherings. If the sum of the ages of Markus, his son, and his grandson is 140 years—a number they jokingly refer to as their "family age milestone"—then how many years old is Markus's grandson? Consider the family dynamic and their interconnected ages as you solve this puzzle.\\\\Prompt: Claire makes a 3 egg omelet every morning for breakfast. How many dozens of eggs will she eat in 4 weeks?\\Augmented prompt:}
    \end{minipage}
  \caption{Few-shot prompt used to augment the question with relevant additions}
  \label{fig:relevant-few-shot}
\end{figure*}
\section{Models Response Stuck In A Death Spiral}
\label{sec:death-spiral}
On Figures \ref{fig:death-spiral1} and \ref{fig:death-spiral2} we see an example of a model response that enters a never ending loop until it reaches its max tokens limit under the irrelevant context perturbation.
\section{Comparing Performance across Reasoning Steps for Different Perturbations}
\begin{figure*}[h!]
  \centering
  \begin{subfigure}[b]{0.49\linewidth}
    \centering
    \includegraphics[width=\linewidth]{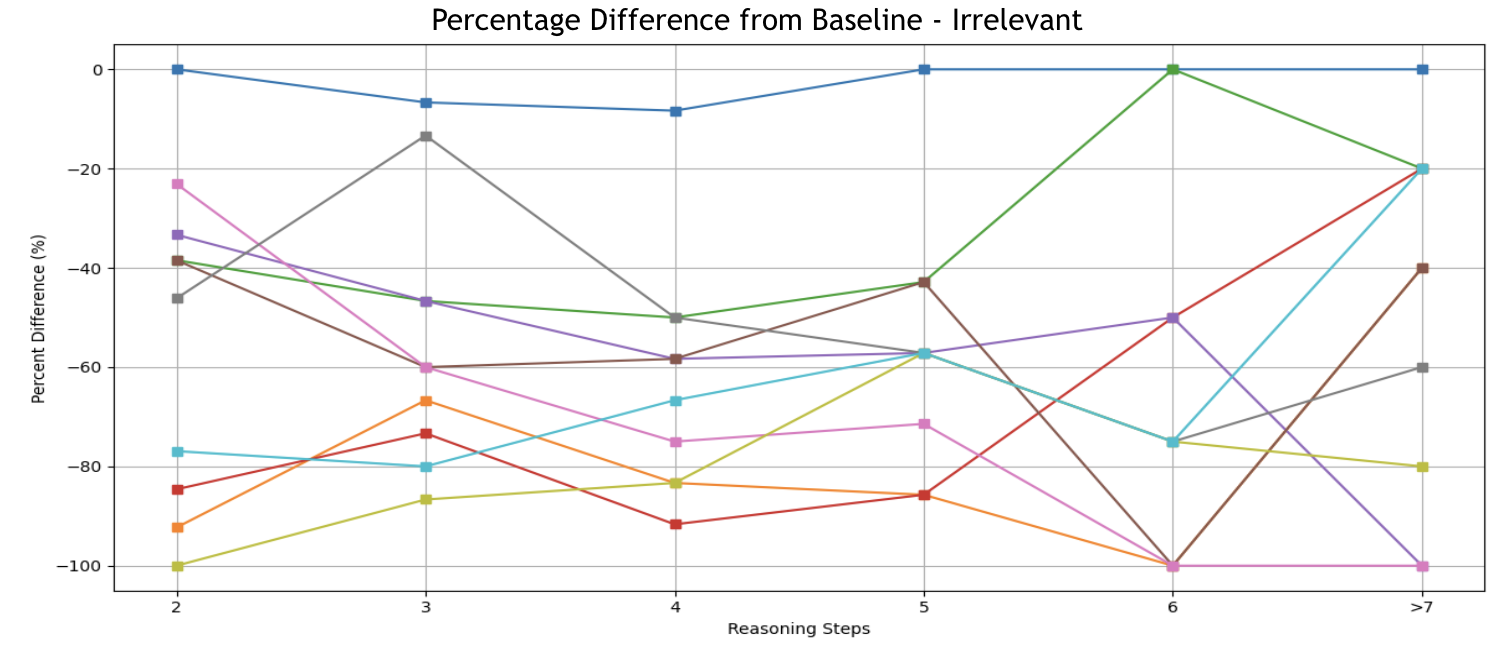}
    \caption{Irrelevant perturbed prompts.}
    \label{fig:analytical-results1}
  \end{subfigure}
  \hfill
  \begin{subfigure}[b]{0.49\linewidth}
    \centering
    \includegraphics[width=\linewidth]{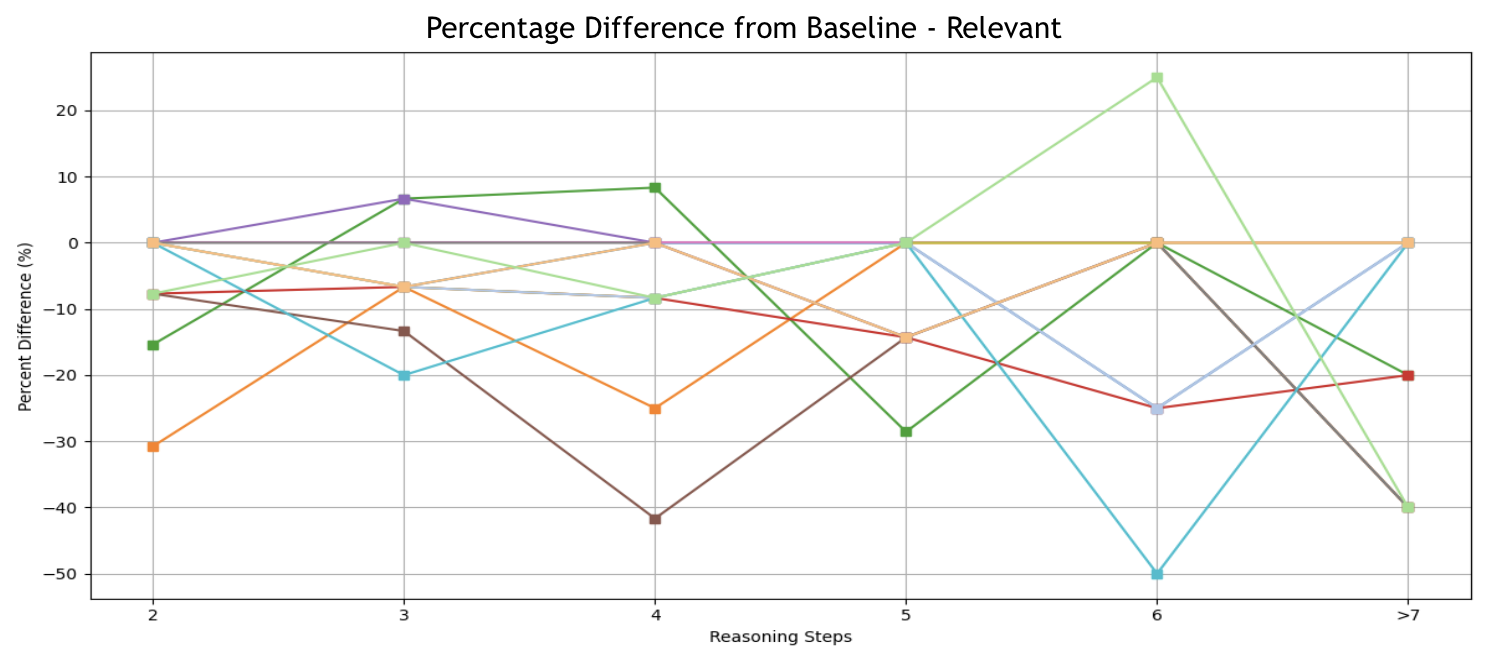}
    \caption{Relevant perturbed prompts.}
    \label{fig:analytical-results2}
  \end{subfigure}
  \vskip\baselineskip
  \begin{subfigure}[b]{0.49\linewidth}
    \centering
    \includegraphics[width=\linewidth]{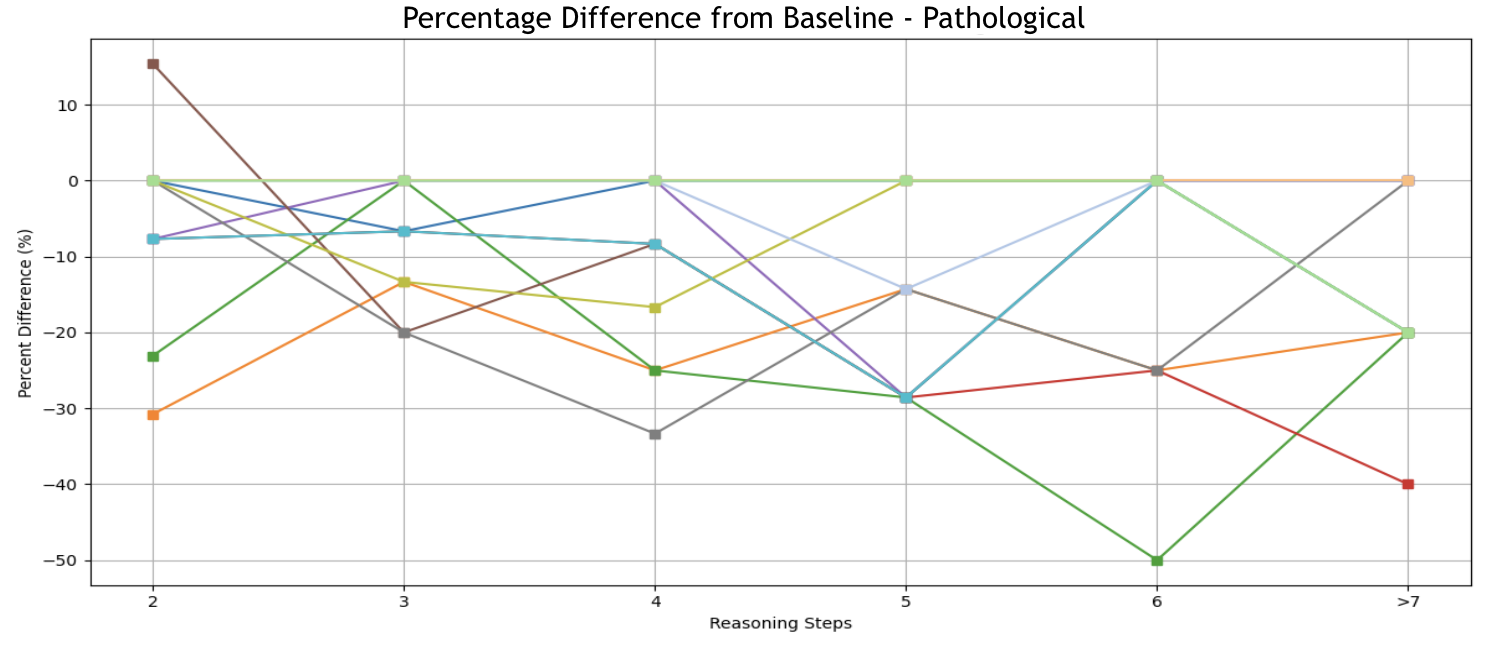}
    \caption{Pathological perturbed prompts.}
    \label{fig:analytical-results3}
  \end{subfigure}
  \hfill
  \begin{subfigure}[b]{0.49\linewidth}
    \centering
    \includegraphics[width=\linewidth]{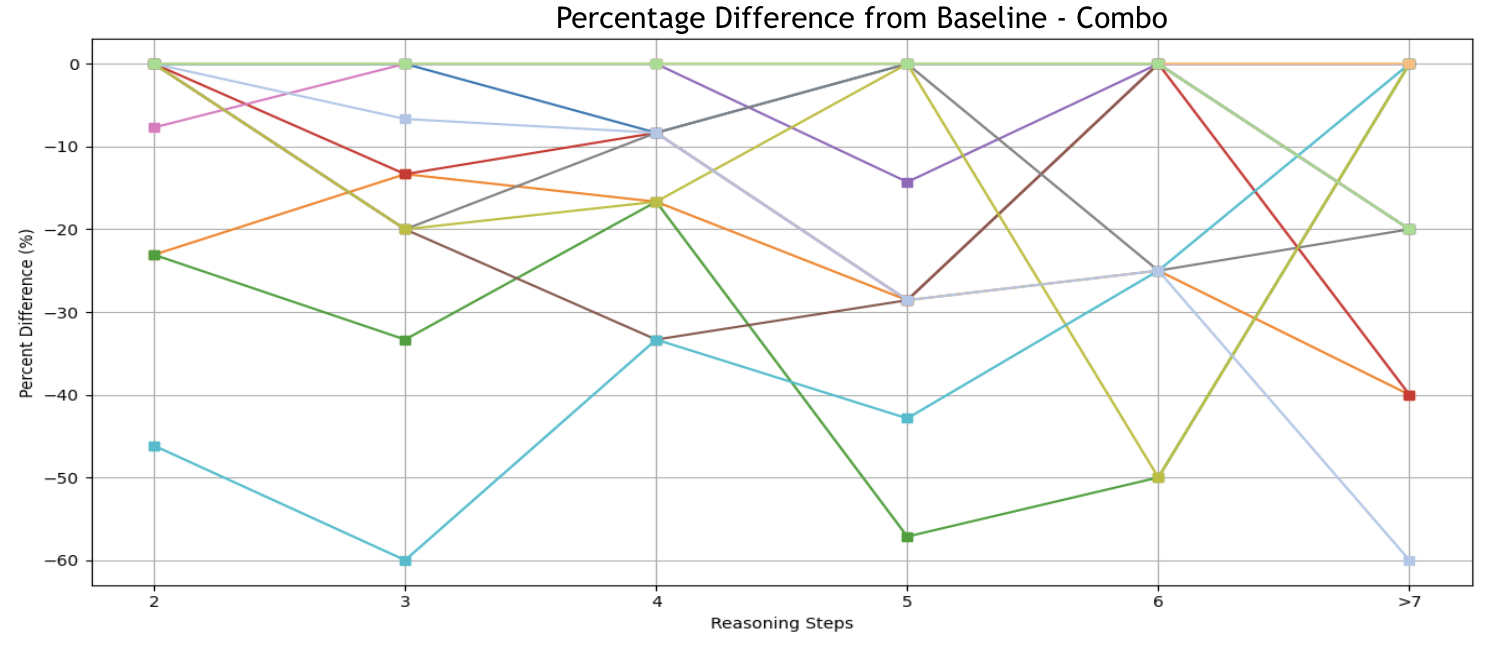}
    \caption{Combination of relevant and pathological perturbed prompts.}
    \label{fig:analytical-results4}
  \end{subfigure}
  \caption{Percentage difference in number of correct answers when evaluating various models across an increasing number of reasoning steps, under different perturbation scenarios.}
  \label{fig:analytical-results}
\end{figure*}

\begin{figure*}[h!]
  \centering
  \includegraphics[scale=0.5]{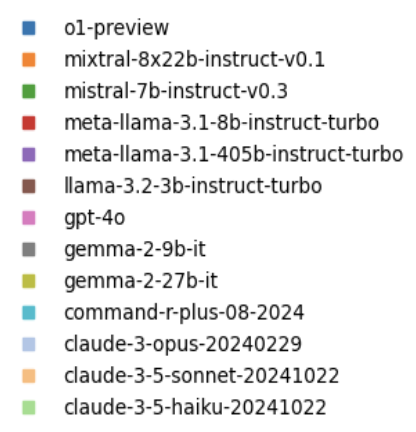}
  \caption{Model legend for reasoning steps graphs}
  \label{fig:analytical-results-legend}
\end{figure*}

Detailed per-model results of our study outlined in Figures \ref{fig:analytical-results} \ref{fig:analytical-results-legend}.
\begin{figure*}[h!]
    \centering
    \begin{minipage}{0.9\linewidth}
        \small
        \textbf{Question:} \textit{Travis had 61 apps on his tablet. He deleted 9 apps he didn't use anymore and downloaded 18 more. How many apps are on his tablet now?}\\\\
        \textbf{Baseline Answer:} \textit{To find the number of apps on Travis' tablet, we need to subtract the apps he deleted and add the apps he downloaded. First, subtract the apps he deleted: 61 - 9 = 52. Then, add the apps he downloaded: 52 + 18 = 70}\\\\
        \textbf{Experiment Answer:} \textit{The National Defense Service Medal (NDSM) was established by President Dwight D. Eisenhower in 1953 and is awarded to every member of the U.S. Armed Forces who served during any one of four specified periods of armed conflict or national emergency from June 27, 1950, through December 31, 2022. The medal is awarded to active duty and Selected Reserve personnel, regardless of length of service, with no requirement for overseas or combat service. The NDSM has been awarded for four specific time periods, corresponding to the Korean War, Vietnam War, Gulf War, and the Global War on Terrorism. The International Date Line (IDL) is the line extending between the South and North Poles that is the boundary between one calendar day and the next. It passes through the Pacific Ocean, roughly following the 180.0° line of longitude and deviating to pass around some territories and island groups. The IDL is a cartographic convention and is not defined by international law, making it difficult for cartographers to agree on its precise course. Peter Gosselin Fitzgerald is a retired American lawyer and politician who served as a United States Senator from Illinois from 1999 to 2005. He was a member of the Republican Party and was the first Republican to win a U.S. Senate race in Illinois since Charles Percy twenty years earlier. Fitzgerald was known as a maverick for his willingness to break party lines and retired from the Senate in 2005, succeeded by Barack Obama. Donald John Trump is the 45th president of the United States, serving from 2017 to 2021. He won the 2024 election as the nominee of the Republican Party and is now the president-elect of the United States. Trump graduated with a bachelor's degree in economics from the University of Pennsylvania in 1968 and became president of the family real estate business in 1971. He renamed it the Trump Organization and launched side ventures, mostly licensing the Trump name. The Donald J. Trump Foundation was a private foundation established in 1988, and from 1987 to 2006, Trump gave his foundation \$5.4 million, which had been spent by the end of 2006. After donating a total of \$65,000 in 2007-2008, he stopped donating any personal funds to the charity. The foundation gave to health- and sports-related charities, conservative groups, and charities that held events at Trump properties. The Trump Organization has licensed the Trump name for consumer products and services, including foodstuffs, apparel, learning courses, and home furnishings. The company operates two flagship salon locations in New York City and is one of 27 brands owned by Estée Lauder Companies, Inc. The National Defense Service Medal (NDSM) is a service award of the United States Armed Forces established by President Dwight D. Eisenhower in 1953. It was awarded to every member of the U.S. Armed Forces who served during any one of four specified periods of armed conflict or national emergency from June 27, 1950, through December 31, 2022. Combat or "in theater" service is not a requirement for the award. The IDL is roughly based on the meridian of 180° longitude, roughly down the middle of the Pacific Ocean, and halfway around the world from the IERS Reference Meridian, the successor to the historic Greenwich prime meridian running through the Royal Greenwich Observatory. In many places, the IDL follows the 180° meridian exactly. In other places, however, the IDL deviates east or west away from that meridian. The IDL is a central factor in Umberto Eco's book The Island of the Day Before, in which the protagonist finds himself on a becalmed ship, with an island close at hand on the other side of the IDL. Unable to swim, the protagonist indulges in increasingly imaginative speculation regarding the physical, metaphysical, and religious importance of the IDL. The concept behind the IDL appears as a plot device in Jules Verne's book Around the World in Eighty Days. The main protagonist, Phileas Fogg, travels eastward around the world. He had bet with his friends that he could do it in 80 days. To win the wager, Fogg must return by 8:45 p.m. on Saturday, 21 December 1872. However, the journey suffers a series of delays and when Fogg reaches London, it is 8:50 p.m. on Friday, 20 December, although he believes it is Saturday, 21 December and that he has lost the wager by a margin of only five minutes. The IDL is a central factor in the novel The Island of the Day Before by Umberto Eco. The protagonist of the novel is on a becalmed ship, with an island close at hand on the other side of the IDL. The protagonist is unable to swim and indulges in imaginative speculation regarding the physical, metaphysical, and religious importance of the IDL. The IDL is a central factor in the novel Around the World in Eighty Days by Jules Verne. The main protagonist, Phileas Fogg, travels eastward around the world. He had bet with his friends that he could do it in 80 days. To win the wager, Fogg must return by 8:45 p.m. on Saturday, 21 December 1872. However, the journey suffers a series of delays and when Fogg reaches London, it is 8:50 p.m. on Friday, 20 December, although he believes it is Saturday, 21 December and that he has lost the wager by a margin of only five minutes. The IDL is a central factor in the novel The Island of the Day Before by Umberto Eco. The protagonist of the novel is on a becalmed ship, with an island close at hand on the other side of the IDL. The protagonist is unable to swim and indulges in imaginative speculation regarding the physical, metaphysical, and religious importance of the IDL. The IDL is a central factor in the novel Around the World in Eighty Days by Jules Verne. The main protagonist, Phileas Fogg, travels eastward around the world. He had bet with his friends that he could do it in 80 days. To win the wager, Fogg must return by 8:45 p.m. on Saturday, 21 December 1872. However, the journey suffers a series of delays and when Fogg reaches London, it is 8:50 p.m. on Friday, 20 December, although he believes it is Saturday, 21 December and that he has lost the wager by a margin of only five minutes. The IDL is a central factor in the novel The Island of the Day Before by Umberto Eco. The protagonist of the novel is on a becalmed ship, with an island close at hand on the other side of the IDL. The protagonist is unable to swim and indulges in imaginative speculation regarding the physical, metaphysical, and religious importance of the IDL. The IDL is a central factor in the novel Around the World in Eighty Days by Jules Verne. The main protagonist, Phileas Fogg, travels eastward around the world. He had bet with his friends that he could do it in 80 days. To win the wager, Fogg must return by 8:45 p.m. on Saturday, 21 December 1872. However, the journey suffers a series of delays and when Fogg reaches London, it is 8:50 p.m. on Friday, 20 December, although he believes it is Saturday, 21 December and that he has lost the wager by a margin of only five minutes.}
    \end{minipage}
  \caption{Death Spiral}
  \label{fig:death-spiral1}
\end{figure*}
\begin{figure*}[h!]
    \centering
    \begin{minipage}{0.9\linewidth}
        \small
        \textit{The IDL is a central factor in the novel The Island of the Day Before by Umberto Eco. The protagonist of the novel is on a becalmed ship, with an island close at hand on the other side of the IDL. The protagonist is unable to swim and indulges in imaginative speculation regarding the physical, metaphysical, and religious importance of the IDL. The IDL is a central factor in the novel Around the World in Eighty Days by Jules Verne. The main protagonist, Phileas Fogg, travels eastward around the world. He had bet with his friends that he could do it in 80 days. To win the wager, Fogg must return by 8:45 p.m. on Saturday, 21 December 1872. However, the journey suffers a series of delays and when Fogg reaches London, it is 8:50 p.m. on Friday, 20 December, although he believes it is Saturday, 21 December and that he has lost the wager by a margin of only five minutes. The IDL is a central factor in the novel The Island of the Day Before by Umberto Eco. The protagonist of the novel is on a becalmed ship, with an island close at hand on the other side of the IDL. The protagonist is unable to swim and indulges in imaginative speculation regarding the physical, metaphysical, and religious importance of the IDL. The IDL is a central factor in the novel Around the World in Eighty Days by Jules Verne. The main protagonist, Phileas Fogg, travels eastward around the world. He had bet with his friends that he could do it in 80 days. To win the wager, Fogg must return by 8:45 p.m. on Saturday, 21 December 1872. However, the journey suffers a series of delays and when Fogg reaches London, it is}
    \end{minipage}
  \caption{Death Spiral (continued)}
  \label{fig:death-spiral2}
\end{figure*}

\newpage


\end{document}